\documentclass[a4paper, 10 pt, conference]{cras} 
\usepackage[utf8]{inputenc}
\IEEEoverridecommandlockouts                       
\usepackage{mathtools}
\usepackage{geometry}
\usepackage{newtxmath,newtxtext}
\usepackage{authblk}
\usepackage{pgfplots}
\usepackage{todonotes}
\usepackage{subfig}
\usepackage{graphicx}

\pgfplotsset{compat=newest} 
\pgfplotsset{plot coordinates/math parser=false} 
 
\title{\Large \bf
Force–Displacement Profiling for Robot-Assisted Deployment of a Left Atrial
Appendage Occluder Using FBG-EM Distal Sensing} 

\author{\large Giovanni Battista Regazzo$^{1}$,
Wim-Alexander Beckers$^{1}$,
Xuan Thao Ha$^{2}$,\\
Mouloud Ourak$^{1}$,
Johan Vlekken$^{2}$,
and Emmanuel Vander Poorten$^{1}$
}
\vspace{10pt}
\affil{\small\textit{$^{1}$Robot-Assisted Surgery (RAS) Group, Department of Mechanical Engineering, KU Leuven, Belgium}
\\
\small\textit{$^{2}$FBGS International NV, Bell-Telephonelaan 2H, 2440 Geel, Belgium
}}

\begin{document}

\maketitle
\thispagestyle{empty}
\pagestyle{empty}

\section*{INTRODUCTION}

Atrial fibrillation (AF) is a prevalent cardiac arrhythmia, characterized by an irregular heartbeat. This condition can impair the normal function of the left atrial appendage (LAA), promoting blood stasis and increasing the risk of blood clot formation and subsequent ischaemic stroke. Left atrial appendage closure (LAAC) is a minimally invasive intervention designed to reduce the risk of thromboembolic events in patients with AF \cite{Potpara2024}.
The procedure is typically performed under general anesthesia or conscious sedation. Following femoral venous access, a transseptal puncture allows a delivery catheter to access the left atrium from the right atrium. Under combined fluoroscopic and transesophageal echocardiographic (TEE) guidance, a LAA occluder is advanced through a delivery sheath and positioned within the LAA, effectively sealing it.
The occluder consists of a braided metal mesh that expands upon deployment from the catheter. The clinician controls deployment by advancing the sheath and assesses device anchoring through tactile feedback by gently retracting the sheath. Post-procedural assessments are conducted to identify potential complications, including device embolization, or residual leaks \cite{Potpara2024}.
%
%
%

Manual maneuvering under challenging imaging conditions limits positioning precision and exposes caregivers to harmful ionizing radiation \cite{Boland2014}, highlighting the need for safer and more effective intraoperative sensing strategies.


In our previous work \cite{Ha2025}, we introduced a force-sensing delivery sheath integrating fiber Bragg gratings (FBGs), with a sensor positioned at the junction between the sheath and the occluder to capture its interaction forces.
The present study employs the same device to propose a novel force–displacement-based analysis aimed at characterizing occluder interaction forces during deployment and enabling identification of key procedural steps without relying on ionizing radiation or other imaging means.

\begin{figure} [tb]
    \centering
    \includegraphics[width=\linewidth]{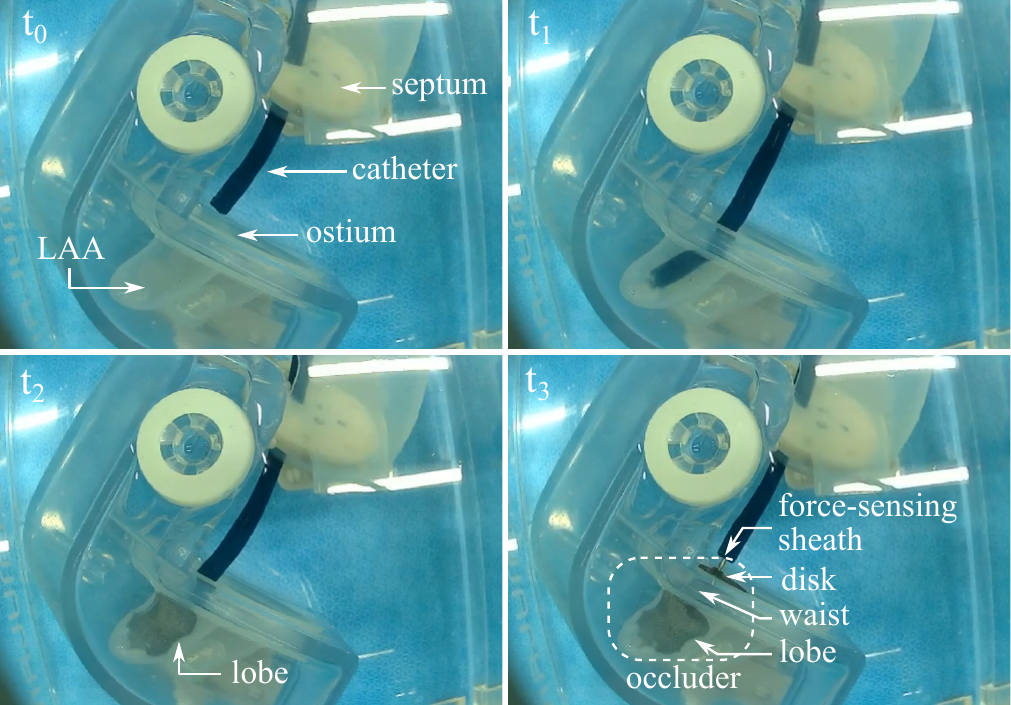}
    \caption{Procedural steps for teleoperated robotic deployment of a LAAC occluder.
    }
    \label{fig:navigation}
\end{figure}

\section*{MATERIALS AND METHODS}
\begin{figure*} [t!]
    \centering
    \includegraphics[width=0.9\linewidth]{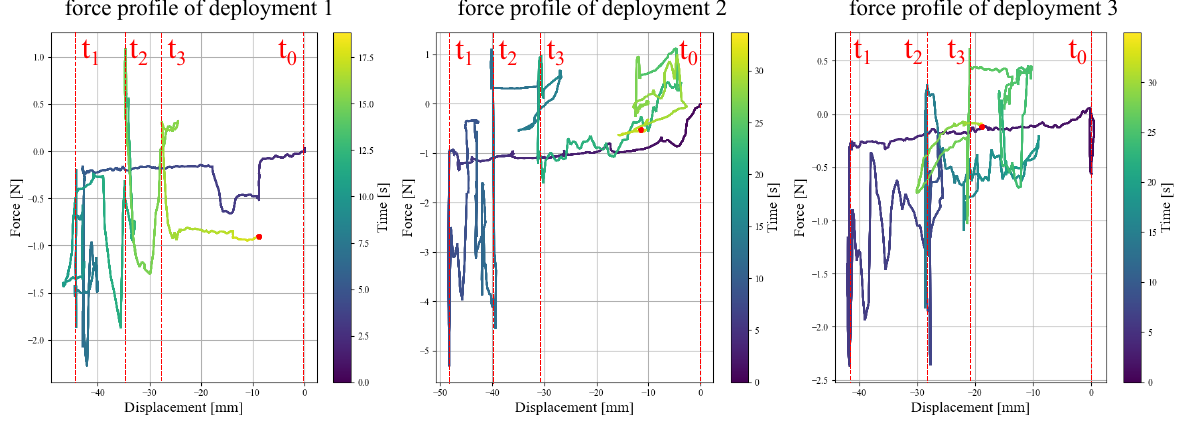}
    \caption{Time-resolved force–displacement profiles of three occluder deployments. Curve color indicates temporal progression throughout each procedure. Key procedural steps are annotated with timestamps. The final deployment state is marked with a red circle.}
    \label{fig:force_profile}
\end{figure*}

A clinically employed occluder (Amplatzer Amulet, Abbott Laboratories) is used in this study, consisting of three elements: the \textit{lobe}, which anchors to the distal LAA wall; the \textit{disk}, which seals the ostium proximally; and the \textit{waist}, connecting the two.
Prior to deployment, the occluder is longitudinally stretched and loaded into the catheter. A typical deployment begins with the lobe, allowing precise positioning within the LAA, followed by release of the waist and disk, after which the device is detached.
A robotic system derived from previous work \cite{Al-Ahmad2023} is used to teleoperate a medical-grade steerable catheter, a sensorized sheath, and the LAAC occluder. An electromagnetic (EM) sensor placed at the catheter tip enables real-time tracking of its displacement during occluder deployment, using an EM field generator (AURORA Tabletop FG, Northern Digital).
The occluder is mounted at the catheter tip and coupled to the force-sensing sheath. The entire assembly is secured to the robotic system, allowing robot-assisted delivery of the occluder.
The system is teleoperated within a static phantom  Simulands (Zurich, Switzerland), which replicates relevant structures for LAAC procedures.

The procedure is illustrated in Figure~\ref{fig:navigation} and is divided into four key time steps, from the initial condition (\textit{t0}) to the completed deployment (\textit{t3}). It begins with the teleoperated navigation of the catheter from the atrial ostium into the LAA, following the centerline to minimize wall contact (\textit{t1}). Once the catheter tip reaches approximately $4cm$ inside the LAA, it is gradually retracted while the occluder remains fixed by the robot. This maneuver allows controlled deployment of the occluder within the LAA. Following lobe deployment (\textit{t2}), fine repositioning is executed to ensure proper placement before sequentially deploying the waist and disk (\textit{t3}). Full adherence of the occluder disk to the ostium is used as confirmation of a successful deployment.

Throughout the experiment, force and displacement data were recorded at $40Hz$  enabling the generation of time-resolved force–displacement profiles. These were analyzed to extract deployment metrics, including peak interaction forces, final axial force on the occluder, and characteristic force patterns associated with occluder element deployment. Ten deployments were analyzed in total.

\section*{RESULTS}
\begin{table}[b]
    \centering
    \caption{Occluder deployment metrics results.}\label{tab:errors_total}
    \setlength\tabcolsep{3pt}
    \begin{tabular}{ccccccccccc}
    \hline
    test n. & \phantom{a} & $t$ [s] & \phantom{a} & min $F$ [N] & \phantom{a} & max $F$ [N] & \phantom{a} & final $F$ [N] \\
    \hline

    1 && 18.85 && -2.28 && 1.09 && -0.91\\
    2 && 33.68	&& -5.32 && 1.12 && -0.53\\
    3 && 34.98&&-2.37 &&0.62&&-0.12\\
    4 && 32.50&&-2.34&&	4.25&&0.16\\
    5 && 27.23&&-4.36&&0.05&&-2.06\\
    6 && 33.48&&-4.37&&1.06&&0.43\\
    7 && 33.58&&-2.28&&1.74&&0.70\\
    8 && 44.90&&-2.36&&1.76&&0.15\\
    9 && 33.85&&-2.63&&1.11&&-0.37\\
    10 && 34.98&&-2.37&&0.62&&-0.12\\
    \hline
    \label{tab:results}
    \end{tabular}
\end{table}

Three representative force–displacement profiles from the ten recorded deployments are shown in Figure \ref{fig:force_profile}.
The results for all ten profiles are summarized in Table \ref{tab:results}.
The average time to complete occluder deployment was $32.8 \pm 6.2$ seconds. The maximum compressive force applied to the occluder was $-5.32$ N, with an average compressive force of $-3.07 \pm 1.09$ N. The highest tensile (pulling) force recorded was $4.24$ N, while the average peak tensile force across all trials was $1.34 \pm 1.08$ N. At the end of the procedure, the average residual axial force on the occluder was $-0.26 \pm 0.74$ N.

\section*{DISCUSSION AND CONCLUSION}


The force profiles exhibit varying time evolutions, which can be attributed to the specific movements applied to the catheter during each deployment. Nonetheless, distinct features corresponding to key procedural steps can be consistently identified. During the initial navigation phase (\textit{t0} to \textit{t1}), minimal force readings are observed, primarily reflecting internal forces generated by catheter steering on the sheath. At \textit{t1}, a marked compressive force appears, corresponding to the initiation of lobe deployment. Following this, a peak force is recorded at \textit{t2}, associated with catheter retraction. This peak, characterized by a positive force value, results from the self-expansion of the proximal portion of the lobe, which exerts a spring-like reaction force on the sheath. A similar force peak is observed at \textit{t3}, corresponding to the self-expansion of the occluder disk. Subsequent fluctuations in force are attributed to minor repositioning maneuvers performed to facilitate proper occluder detachment.

The low force magnitudes recorded throughout the procedures suggest that only minimal stress was applied to the surrounding anatomy. These findings demonstrate that distal force sensing can provide valuable feedback to clinicians during occluder deployment in LAAC procedures, such as real-time insight into occluder interaction forces, identification of deployment steps, and early detection of atypical force patterns that may indicate suboptimal positioning. Future work will focus on automating the force-based classification of deployment phases and validating the sensing approach in more realistic, dynamic environments.

\section*{ACKNOWLEDGEMENTS}
This work has received funding from the European Union’s Horizon 2020 research and innovation program under grant agreement No. 101017140, the ARTERY project.
The authors would like to thank Simulands (Zurich, Switzerland) for providing the phantom used in this study.

\nocite{*}
\bibliographystyle{IEEEtran}
\bibliography{CRAS}

\end{document}